\documentclass[10pt,letterpaper]{article}

\usepackage{cogsci}

\cogscifinalcopy

\usepackage{pslatex}
\usepackage{apacite}
\usepackage{float}

\usepackage{booktabs}
\usepackage{graphicx}
\usepackage{gensymb}
\usepackage{amsmath}
\usepackage{amsfonts}
\usepackage{algorithm} 
\usepackage{algpseudocode} 
\usepackage{bbm}

\title{Developmental Curiosity and Social Interaction in Virtual Agents}

\author{Chris Doyle\textsuperscript{1}, Sarah Shader\textsuperscript{2}, Michelle Lau\textsuperscript{3}, Megumi Sano\textsuperscript{4}, Daniel L. K. Yamins\textsuperscript{4,5} \& Nick Haber\textsuperscript{1} \\ \\ \textsuperscript{1}Graduate School of Education, Stanford University, 
\textsuperscript{3}Stanford University \\
\textsuperscript{4}Department of Computer Science, Stanford University,
\textsuperscript{5}Department of Psychology, Stanford University \\
\texttt{\{crd, mblau, megsano, yamins, nhaber\}@stanford.edu} \\ \\
\textsuperscript{2}Department of Computer Science, Pomona College \\
\texttt{sdsb2018@mymail.pomona.edu} }
\begin{document}

\maketitle

\begin{abstract}
Infants explore their complex physical and social environment in an organized way. To gain insight into what intrinsic motivations may help structure this exploration, we create a virtual infant agent and place it in a developmentally-inspired 3D environment with no external rewards. The environment has a virtual caregiver agent with the capability to interact contingently with the infant agent in ways that resemble play. We test intrinsic reward functions that are similar to motivations that have been proposed to drive exploration in humans: surprise, uncertainty, novelty, and learning progress. These generic reward functions lead the infant agent to explore its environment and discover the contingencies that are embedded into the caregiver agent. The reward functions that are proxies for novelty and uncertainty are the most successful in generating diverse experiences and activating the environment contingencies. We also find that learning a world model in the presence of an attentive caregiver helps the infant agent learn how to predict scenarios with challenging social and physical dynamics. Taken together, our findings provide insight into how curiosity-like intrinsic rewards and contingent social interaction lead to dynamic social behavior and the creation of a robust predictive world model.

\end{abstract}

\section{Introduction}

Infants are born into a complex set of social and physical phenomena. At the center of their world are caregivers who smile at them, change their diapers, point at things, and sing songs, and around them there are bouncing balls, falling block towers, and spinning tops. Infants must figure out how to control their bodies and learn how the world responds to their actions. Infants' exploration is not random, they explore their world in a structured way \cite{gopnik1999scientist}. 

Over time, children develop an understanding of their world. Infants are sensitive to social contingency, the reactions of others to their actions \cite{nadel1999expectancies} and the level of responsiveness of a partner \cite{bigelow2006two}. Infants have expectations about how people will respond to their actions \cite{tronick1978infant} and how objects will behave \cite{stahl2015observing}. 

A compelling hypothesis is that the motivation to explore may be linked to a desire to improve the accuracy of predictions about the world. Working to improve these predictions (the agent's ``world model") can create a self-generated learning curriculum, through a cycle of evaluating deficiencies in the model, seeking out information, updating the model, and gaining new capabilities \cite{schmidhuber2010formal}. Researchers have found evidence that suggests violations of expectation catalyze learning \cite{stahl2015observing}, and that learning progress is an important component for task selection \cite{ten2021humans}. Children appear sensitive to the discriminability of hypotheses and explore longer when hypotheses are harder to distinguish \cite{siegel2021children}. Stimulus novelty may also play a role in curiosity-driven exploration \cite{poli2022contributions}. Children can effectively explore diverse scenarios, including both physical and social phenomena. Intrinsic reward functions implemented in reinforcement learning contexts are more fragile and can be susceptible to white-noise fixation \cite{oudeyer2007intrinsic, schmidhuber2010formal, pathak2017curiosity}, or may not lead to meaningful behavior diversity.

\begin{figure}[tp]
\begin{center}
    \includegraphics[width=\columnwidth]{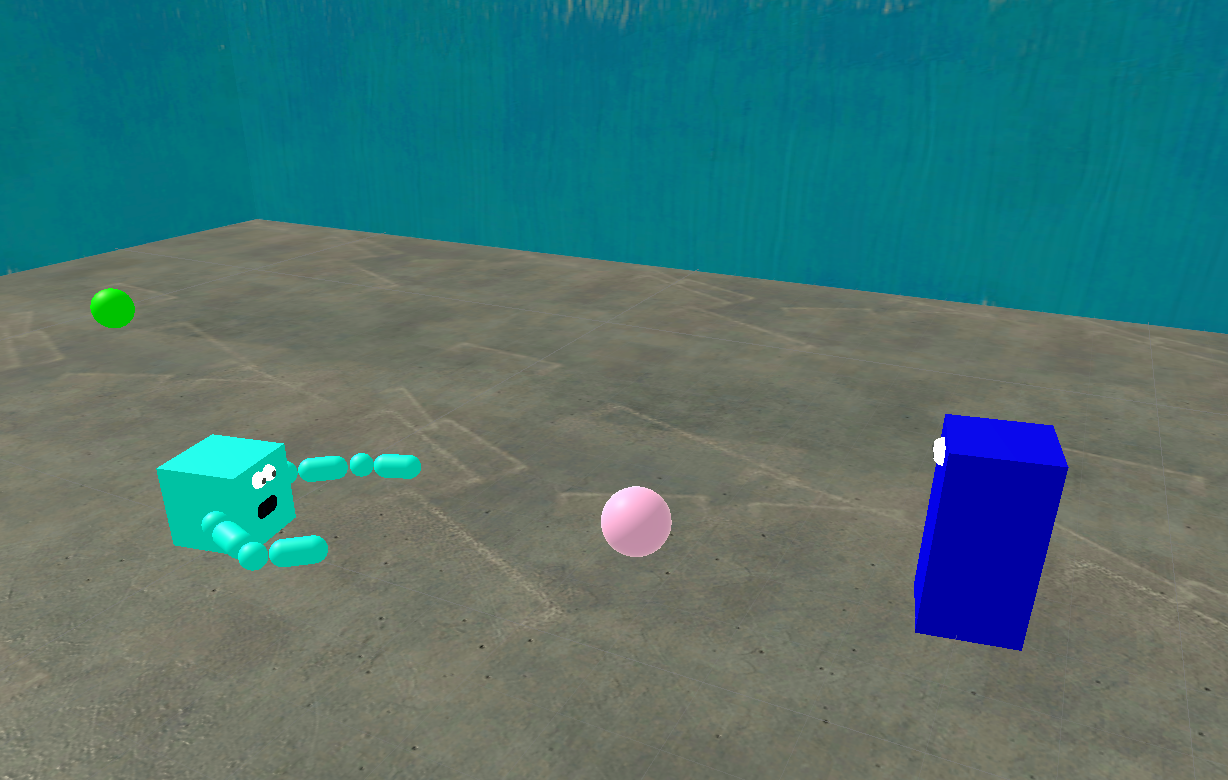}
    \vspace{-20pt}
\end{center}
\caption{The environment is a room that contains an infant agent (teal), caregiver agent (dark blue), and two movable balls (pink and green) } 
\label{env-picture}
\vspace{-20pt}
\end{figure}

\begin{figure*}[tp]
\centering
\includegraphics[width=\textwidth]{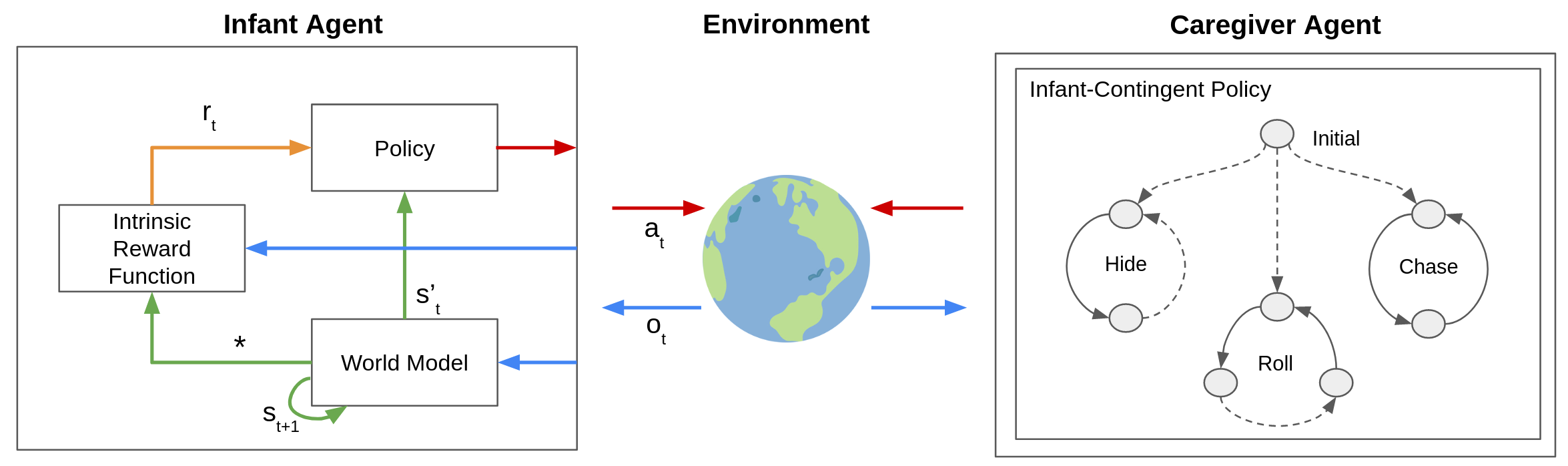}
    \vspace{-25pt}
  \caption{The infant agent's world model, reward function, and policy interact to drive infant actions over time. The intrinsic reward function takes different inputs depending on the choice of function. The state diagram (right) outlines the caregiver's static policy. Dotted lines indicate transitions that depend on the infant state, and solid lines indicate those that do not. The top circle is the starting state, where the caregiver waits for the infant to point, and the three branches are unlocked by the infant pointing at one of the objects in the room.}
  \label{infant-architecture}
      \vspace{-15pt}
\end{figure*}

Previous work showed that intrinsic rewards lead to exploration in a physical context \cite{haber2018learning} and a preference for viewing animate objects in a protosocial context \cite{kim2020active}, but it did not include complex social contingencies or a sophisticated embodiment for the agent. We extend the work in these directions to evaluate if a curiosity-like intrinsic reward function can generate social behavior in virtual agents and to examine the effect of contingency on how a virtual agent learns social and physical dynamics.

We summarize our contributions as follows.
\begin{itemize}

\item We introduce a developmentally-inspired virtual 3D environment with an embodied infant agent and a caregiver agent that can engage in complex, contingent, social behaviors with the infant.

\item We describe an infant agent with an intrinsic reward function inspired by motivations hypothesized to be present in humans. The infant agent learns to pursue intrinsic rewards through reinforcement learning. We show that the agent generates temporally variable, social, play-like behavior within our environment, in the absence of extrinsic reward. When motivated by a novelty reward or an uncertainty reward, the agent builds world models that can make good predictions about experiences beyond their own.

\item We show that a high level of contingency in the caregiver agent corresponds with the infant agent learning to make better predictions about challenging scenarios involving caregiver and object dynamics.
\end{itemize}

\section{Environment}
Our 3D virtual environment is created in Unity and uses the ML-Agents framework \cite{juliani2018unity}. We use episodes of 2,000 timesteps over 200 in-environment seconds. At the end of an episode, the environment is reset to its starting state.

The setting is a closed room containing two ball objects, a caregiver agent, and an infant agent, pictured in Figure \ref{env-picture}. The Unity physics engine allows the objects to respond to forces applied to them by the infant's body and arms. The balls can also be picked up and thrown by the caregiver. 

\subsection{Infant}
The infant has two arms with shoulder and elbow joints. The arms can only move in the plane parallel to the floor and are at a height they can collide with the ball. At each timestep, the infant can choose one of 13 actions: do nothing, turn left/right, move forward/back, or rotate any of the four arm joints clockwise/counterclockwise. 

The infant has partial observability: it receives an indicator as to whether each object is in its field of view (120\degree forward), and if the object is in view, its position, orientation, and velocity. It receives proprioceptive information giving the positions and orientations of its arms and its body, and the value of a hit sensor on each arm.

\subsection{Caregiver}
The caregiver agent can move around the room and pick up and throw the balls. It is controlled by a script that begins each episode watching the infant agent and waiting for the infant to ``point" to an object. Pointing is determined by the infant orienting their body toward an object, with an arm pointed straight forward, and holding that position for five timesteps. If the infant points toward an object or the caregiver, a branch of the script is activated. Pointing toward the caregiver activates the ``hide and seek" branch (Hide), pointing toward the pink ball activates the ``roll to infant" branch (Roll), and pointing toward the green ball activates the ``chase the ball" branch (Chase). At the end of an episode, the environment is reset and the caregiver waits for the infant to point again. The high-level state diagram is shown in Figure \ref{infant-architecture}.

\subsubsection{Hide and seek} The caregiver selects a point in the area behind the infant and moves there. When it arrives, it waits for the infant to look in its direction, at which point the caregiver selects a new point to move to behind the infant. 

\subsubsection{Roll to infant} The caregiver retrieves the pink ball, moves a target distance from the infant, then looks at the infant. The caregiver waits for the infant to look in its direction. Once that occurs, the caregiver rolls the ball to the infant and waits for a fixed period before retrieving the ball and starting again.

\subsubsection{Chase the ball} The caregiver continually retrieves the green ball and throws it forward. This cycle causes the ball to be thrown around the room, bouncing off the walls and floors.

\subsubsection{Independent play} If no object is pointed toward, the caregiver will remain looking at the infant for the entire episode.

\section{Infant Agent}
\begin{figure*}[t]
    \includegraphics[width=\textwidth]{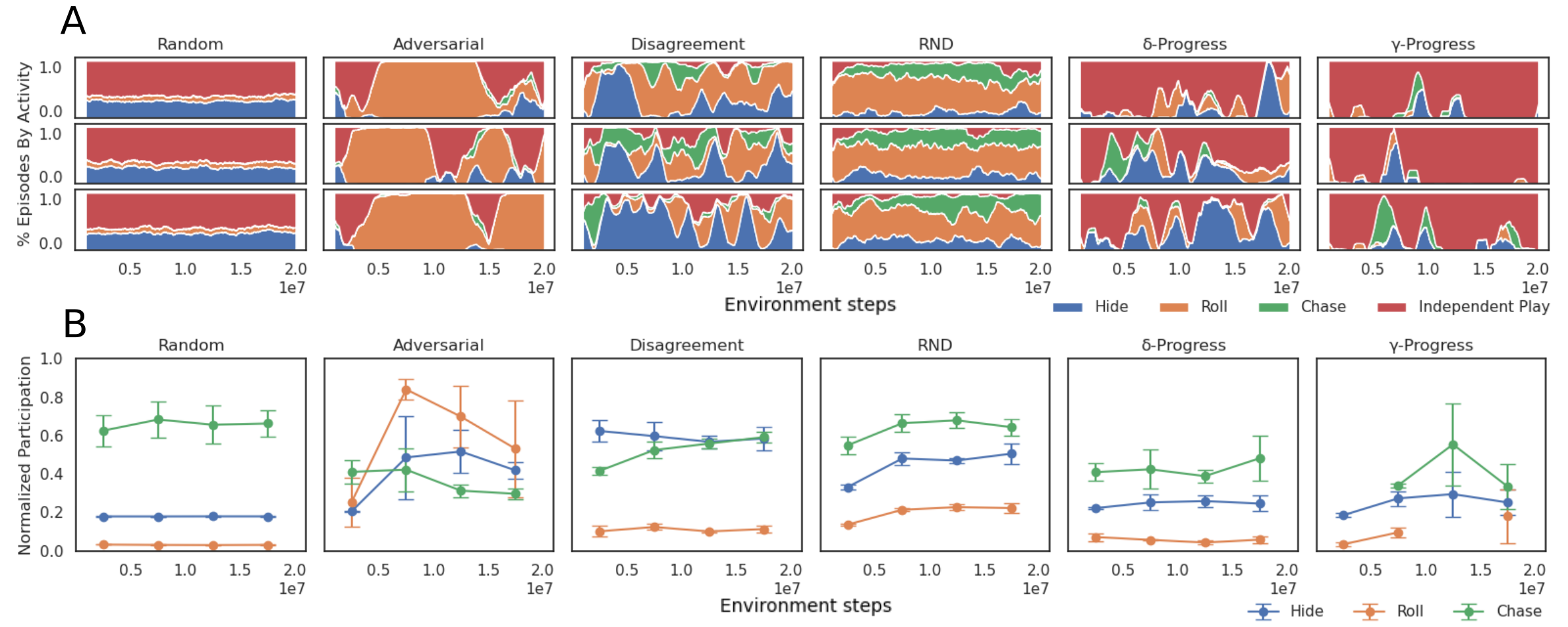}
    \vspace{-25pt}
\caption{A. The proportion of episodes that the infant agent activates one of the contingent behaviors (Hide, Roll, Chase) or if they do not activate a behavior (Independent Play). Three random seeds are shown for each reward function. B. The average participation by behavior is calculated across 5M steps. The standard error is shown across the three random seeds. Values are normalized by dividing them by the 99th percentile of the metric across training.} 
\label{behavior-diversity}
    \vspace{-15pt}
\end{figure*}

The infant agent has three primary components that drive its behavior over time: a world model, an intrinsic reward function, and a policy (Figure \ref{infant-architecture}). The components are intended to be only high level analogies to human capabilities, they do not represent a mechanistic hypothesis. We choose implementations that are computationally tractable and have been successful in various reinforcement learning contexts.

\subsection{World model}
The objective of the infant agent's world model is to predict the next observation given the history of observations and actions. We create a latent dynamics model that attempts to model changes in the underlying the environment state.

\subsubsection{Architecture}  
The world model uses a two-layer LSTM \cite{hochreiter1997long}. In addition to maintaining the hidden states of $h$ and $c$ in the LSTM, the world model maintains a belief state $b$ that contains an estimate of position, orientation, and velocity for all objects in the environment. The $b$ component has a dimensionality equal to that of an observation $o$, less the object-visibility indicator dimensions. The augmentation allows the infant agent to maintain a belief about objects outside its visual frame, a capability that humans have \cite{piaget1952origins}. We refer to the tuple of world model states $(h, c, b)$ as $s$. Changes to $b$ are predicted using an MLP decoder on $h$. Delta prediction for physical quantities has been successful in fully-observed physics prediction \cite{battaglia2016interaction, chang2016compositional} and we adapt it for a partially observed setting.

\begin{algorithm}[t]
	\caption{Agent algorithm} 
	\begin{algorithmic}[1]
        \State \textbf{Input} total episodes $E$, episode length $T$, world model training iterations per episode $M$, batch size $N$, sequence training length $L$, intrinsic reward function $\mathcal{R}$
        \State Initialize replay buffer R = $\emptyset$, parameters for world model $\theta$, policy $\phi$, and intrinsic reward $\psi$       
		\For {$\text{episode}=1,2,\ldots E$}
            \State Initialize belief $b$ and LSTM hidden states $(h, c)$
            \For {$t=1,2, \ldots T$}
                \State Observe $o_t$
                \State Update $s_{t}$ to $s'_{t}$ with information from $o_t$
                \State $a_t \sim \pi_{\phi}( a | s'_{t})$ 
                \State $s_{t+1} \leftarrow f_{\theta}(s'_{t}, a_t)$ 
                \State Take action $a_t$
			\EndFor
            \State Add collected tuples of $(o, a, s)$ to replay buffer $R$
            \State Calculate reward $r_t$ for steps $1...T$ using $\mathcal{R}$
            \State Update $\phi$ with PPO, update $\psi$ as applicable
            \For {$i=1,2,\ldots, M$}
                \State Sample $N$ sequences with length $L$ from $R$
                \State Calculate $\mathcal{L}_{\text{WM}}$ on batch and update $\theta$
            \EndFor
		\EndFor
	\end{algorithmic} 
\end{algorithm}

\subsubsection{Training} 
The world model is supervised on rollouts of length $L = 30$. We use a stored hidden state and burn-in to help prediction accuracy \cite{kapturowski2018recurrent}. The world model is recurrently applied $s_{t+1} \leftarrow f_{\theta}(s_{t}, a_t)$ using the action sequence from the replay buffer. Predicted observations $\hat{o}$ are directly read out of the $b$ state. The world model loss is the square error of the visible components of the observation over the rollout.

\vspace{-15pt}
\begin{equation}
\mathcal{L}(\hat{o}_{1...L}, o_{1...L}) = 
\sum_{i=1}^{L}{ 
    \left( 
        \sum_{j=1}^{\dim{(o_{i})}}
        {
            (\hat{o}_{i,j} - o_{i,j})^2 \mathbbm{1}_\text{visible}(o_{i,j}) 
        } 
    \right) 
}
\end{equation}

\subsection{Intrinsic reward functions}
\subsubsection{Adversarial} A violation of expectation can be framed as the prediction from a world model being significantly different from the observed outcome. This surprise-based intrinsic reward can be formulated as a function of the prediction error \cite{achiam2017surprise, pathak2017curiosity, schmidhuber2010formal}. We use the model loss as the reward.

\subsubsection{Disagreement} Being uncertain about the outcome of an action can be interpreted as there being variance around a prediction of the future. Uncertainty has been formulated as the variance of predictions across an ensemble of trained world models \cite{pathak2019self, sekar2020planning}. We use $K = 10$ models in the ensemble. Because of memory and training time constraints, the recurrent dynamics model is not replicated. Instead, the ensemble members are MLPs that predict the next observation from the state $s$ and action.

\subsubsection{Random Network Distillation (RND)} Novel stimuli can indicate the potential for learning. In environments with a discrete state space, a reward that is a decreasing function of visit counts can be effective in incentivizing exploration \cite{strehl2008analysis}. Although that approach is not directly applicable to continuous spaces, methods such as pseudo-counts \cite{tang2017exploration} and Random Network Distillation \cite{burda2018exploration} can be used.

\subsubsection{Learning Progress} An agent may try to pursue experiences that are likely to improve its understanding of the world. One approach to estimating this is to evaluate recent learning progress on that topic, that is, the magnitude of improvement between a previous world model and the current one. This has been implemented as $\delta$-progress \cite{achiam2017surprise, graves2017automated} and $\gamma$-progress \cite{kim2020active}. The difference between the methods is how the previous world model is defined: $\delta$-progress uses a world model from $\delta$ steps ago and $\gamma$-progress updates the weights of the old model by performing a weighted average of old weights with current weights.

\subsection{Policy learning}
We modify Proximal Policy Optimization (PPO) \cite{schulman2017proximal}, a model-free reinforcement learning algorithm, to have the learned policy be based on the world model state $s$ instead of observations.

\section{Experiment 1: Compare exploration across intrinsic reward functions}
We investigate two questions: what type of behavior diversity arises from different reward functions and which reward functions lead the infant agent to learn a robust world model.

\subsection{Method}
Infant agents are assigned one of the intrinsic reward functions and trained for 20M steps. Three random seeds are run for each reward function. After training, we analyze the behavior and evaluate the infant agent's world model.

\subsubsection{Behavior diversity}
We consider behavior diversity in three ways: state coverage, social contingency activation, and level of contingency participation.

To evaluate state coverage, we independently consider four components of the infant agent's observations: its location within the room, its orientation, its pose, and what objects, animate or inanimate, are visible to it. We calculate the normalized entropy as the entropy of a discretized version of the component relative to a uniform distribution (Table \ref{entropy-infant-state}).

Activating and participating in the social contingencies is particularly important because it allows the infant agent to unlock new parts of the state space. We look at the proportion of episodes where the behavior is activated. For each of the activated behaviors we identified a metric that corresponds to ``participation" in the activities: within the Hide behavior, the number of times the infant finds the caregiver; within the Roll behavior, the number of times the infant hits the ball; within the Chase behavior, the frequency that the infant is looking at the caregiver when the ball is thrown. 

\subsubsection{World model performance evaluation}

We assess the robustness of a world model by evaluating its predictions on trajectories it has not been trained on. We test it on experiences collected by agents with different seeds and different intrinsic reward functions, and on experiences collected by manually programmed agents.

For each agent, we create a set of validation cases from its lifetime experience by uniform sampling 2000 trajectory segments. In a round-robin fashion, we test the world model from each agent against the validation case sets for each other agent, including different seeds and different intrinsic reward functions. We score the model on each validation case set by calculating the average total model loss over a 10-step rollout. 

\subsection{Results}

\begin{figure}[t]
\begin{center}
    \includegraphics[width=\columnwidth]{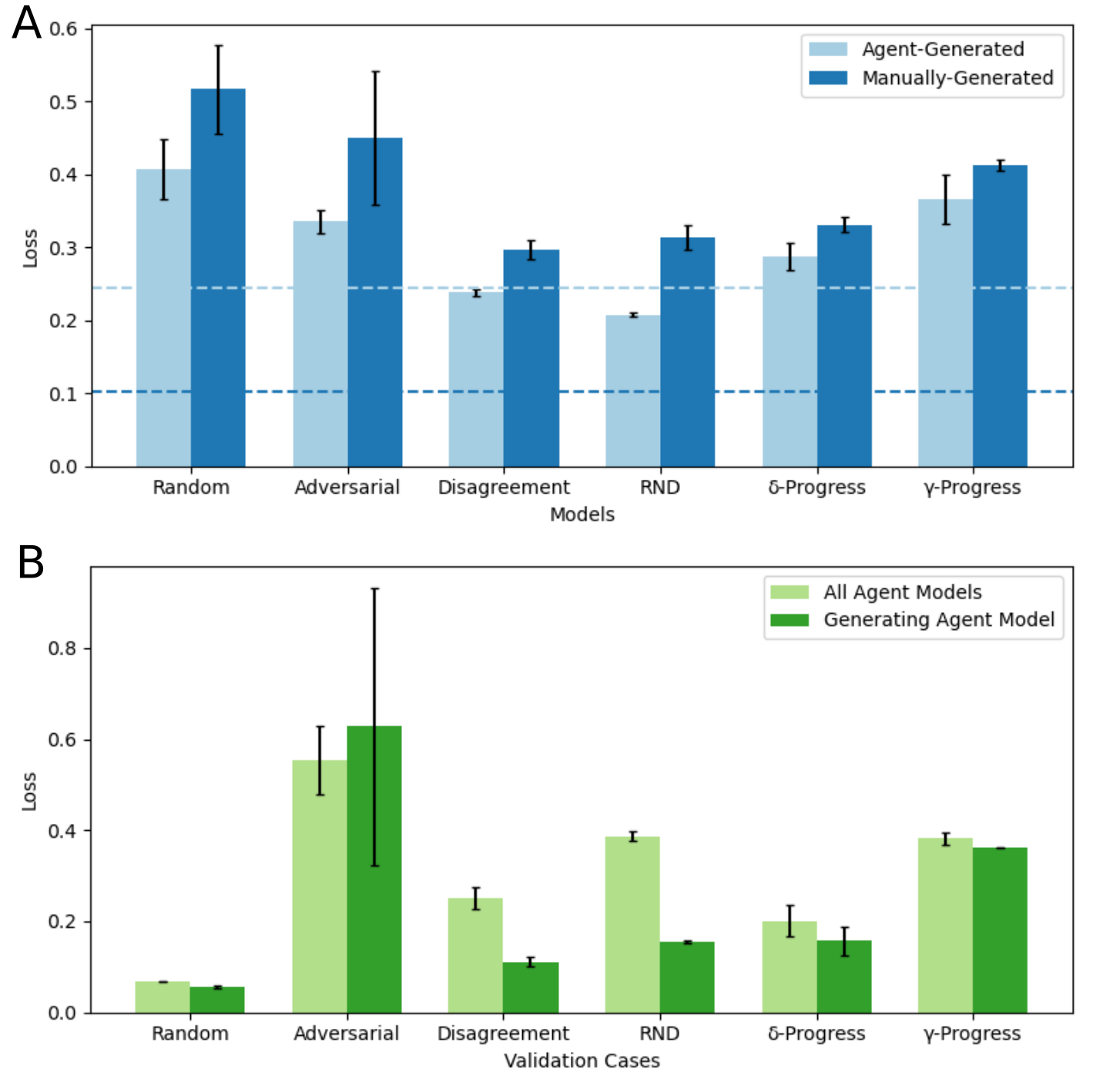}
\end{center}
\vspace{-20pt}

\caption{A. World model loss on validation cases generated from the experience of other agents and validation cases that were created manually. Lower values indicate better accuracy. Agent-Generated cases include validation sets from all seeds of all intrinsic reward functions. The horizontal lines are the average loss if each validation case is predicted using the world model from the agent that generated the data (an estimate of good performance in our model class). B. World model loss on a validation case set. All Agent Models is the average loss on a set across all agents. Generating Agent Model is the loss on the set using the world model from the agent that generated the set. }
\label{model-performance-and-validation}
\vspace{-16pt}
\end{figure}

\begin{figure*}[t]
\begin{center}

    \includegraphics[width=\textwidth]{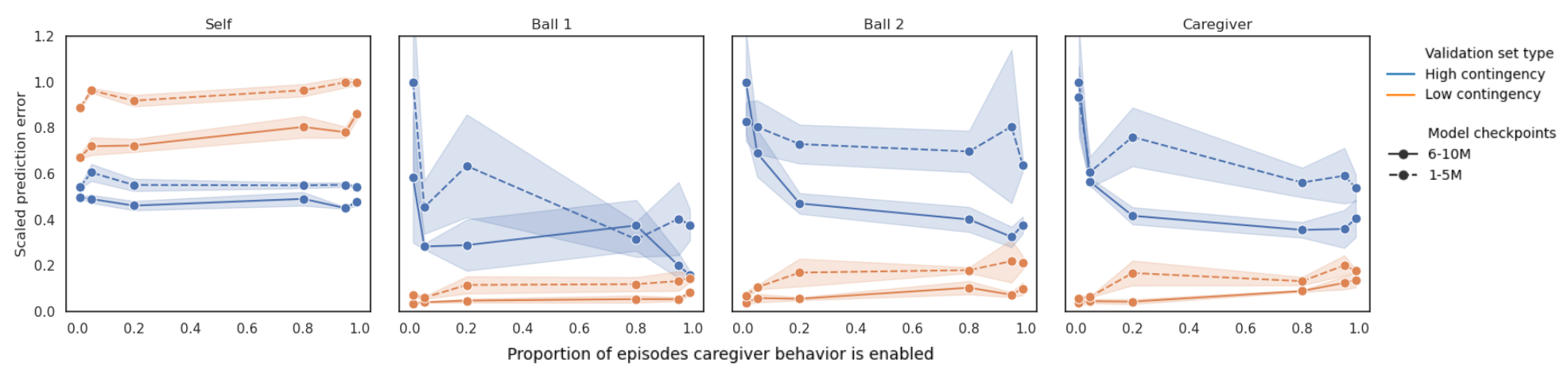}

\end{center}

\vspace{-20pt}

\caption{Prediction error by observation component shown plotted against the proportion of time caregiver contingency is enabled during training. Lower values indicate better accuracy. The error is decomposed into that due to infant orientation, position, and arm configuration ("Self"), Ball 1, Ball 2, and Caregiver predictions. High contingency refers to the set of validation cases created from agents where caregiver contingency was enabled 95\% of the time or more, and low contingency refers to where caregiver contingency was enabled 5\% of the time or less. The results are shown for world model checkpoints taken from 1M to 5M training steps and 6M to 10M training steps and plotted with the standard error across three seeds. } 
\label{disag-loss-by-contingency}
\vspace{-15pt}
\end{figure*}

\subsubsection{Disagreement and RND yield the greatest diversity of experience and acquire the most robust world models} 

\begin{table}[t]
\small
\begin{center} 
\caption{Normalized entropy of infant state components. Calculated out of 100. Mean and standard error over 3 seeds.} 
\label{entropy-infant-state} 

\vspace{-5pt}
\begin{tabular}{lllll} 
\hline
Agent & Location & Orientation & Pose & Attention  \\
\hline
Random & 5 $\pm$ 0 & 56 $\pm$ 0 & 100 $\pm$ 0 & 62 $\pm$ 0  \\
Adversarial & 45 $\pm$ 5 & 79 $\pm$ 2 & 76 $\pm$ 4 & 71 $\pm$ 1 \\
Disagreement & 93 $\pm$ 0 & 100 $\pm$ 0 & 99 $\pm$ 0 & 95 $\pm$ 0 \\
RND & 87 $\pm$ 1 & 98 $\pm$ 0 & 99 $\pm$ 0 & 93 $\pm$ 0 \\
$\delta$-progress & 40 $\pm$ 7 & 88 $\pm$ 1 & 94 $\pm$ 2 & 80 $\pm$ 1   \\
$\gamma$-progress & 38 $\pm$ 4 & 80 $\pm$ 3 & 90 $\pm$ 3 & 74  $\pm$ 1 \\
\hline
\end{tabular} 
\end{center} 
\vspace{-20pt}
\end{table}

Disagreement and RND generated higher entropy (Table \ref{entropy-infant-state}) than all other intrinsic reward functions across the Location, Orientation, and Attention components of state (Location: $p < 0.005$; Orientation: $p < 0.01$; Attention: $p < 0.001$; t-test with fdr-bh correction). They also generate a larger number of total activations (Table \ref{table-activations}) than the random agent, $\delta$-progress, and $\gamma$-progress ($p<0.01$), and RND shows higher total and Chase activations than Adversarial ($p<0.05$). 

This behavior diversity likely contributes to the large spread in Figure \ref{model-performance-and-validation}B between 1) the model loss from other agents' world models on experiences from Disagreement and RND, and 2) the low loss that Disagreement and RND can achieve on its own validation cases. This implies that the cases are predictable by this model class, but that the other models have not learned the dynamics yet.

The world model loss from Disagreement and RND agents (Figure 4A) is significantly lower than Random, Adversarial, and $\gamma$-progress agents when predicting the outcome of experiences that other instrinsically-motivated agents collected ($p<0.05$; t-test with fdr-bh correction and one high-loss outlier removed in Adversarial and $\gamma$-progress). Interestingly, $\delta$-progress also does well on our world model evaluation despite having fewer contingency activations, lower participation, and lower state entropy.

Although RND and Disagreement both generate a high diversity of states, they have different temporal structures in behavior (Figure \ref{behavior-diversity}A). RND has a relative stable split of behavior activations, whereas Disagreement generates varying proportions of behavior activations over time. In general, reward functions seem to generate behavior patterns that are distinguishable from each other and consistent across seeds. 

\subsubsection{Agents using the Adversarial reward function focus on hitting the ball in the Roll behavior}

Agents driven by the adversarial signal frequently activate the Roll behavior. Within the Roll behavior the agent spends most of the time hitting the ball back and forth between its arms. The described behavior contributes to the lower entropy observed in Table \ref{entropy-infant-state} and the high Roll participation in Figure \ref{behavior-diversity}B. 

The intrinsic reward function rewards high loss situations. The agent finds them and validation cases made from the Adversarial agent's experience have the highest average loss across models (Figure \ref{model-performance-and-validation}B). Even with the benefit of the experience, the loss remains high for the world model of the agent that collected it. The agent found situations that are difficult to model accurately with the current world model class.

\begin{table}[t]
\small
\begin{center} 
\caption{Percent of episodes where behavior activation occurs over agent training. Mean and standard error over 3 seeds.} 
\label{table-activations} 
\begin{tabular}{lllll}
\hline
Agent & Hide & Roll & Chase & Total \\
\hline
Random & 30 $\pm$ 0 & 9 $\pm$ 0 & 0 $\pm$ 0 & 39 $\pm$ 0 \\
Adversarial & 7 $\pm$ 2 & 64 $\pm$ 6 & 2 $\pm$ 0 & 73 $\pm$ 4 \\
Disagreement & 38 $\pm$ 5 & 35 $\pm$ 7 & 14 $\pm$ 4 & 87 $\pm$ 3 \\
RND & 15 $\pm$ 2 & 53 $\pm$ 2 & 23 $\pm$ 1 & 91 $\pm$ 1 \\
$\delta$-Progress & 23 $\pm$ 6 & 14 $\pm$ 2 & 4 $\pm$ 2 & 42 $\pm$ 7 \\
$\gamma$-Progress & 7 $\pm$ 1 & 3 $\pm$ 0 & 4 $\pm$ 3 & 14 $\pm$ 3 \\
\hline
\end{tabular}
\end{center} 
\vspace{-20pt}
\end{table}

\section{Experiment 2: Vary the frequency the caregiver responds to the infant agent}

Social interaction is critical for typical development \cite{kaler1994analysis}. We use our simulated environment to analyze how the infant agent's understanding of the world is affected when it interacts with a less attentive caregiver compared with a more attentive one.
\subsection{Method}

Infant agents are trained for 10M steps using the Disagreement reward function, selected based on the high state entropy and low world model loss observed in Experiment 1 on both agent- and manually-generated validation sets. The analysis is limited to one reward function due to computational constraints.
In contrast to Experiment 1, the caregiver response to the infant ``pointing" is stochastic. A flag is set with some probability at the beginning of each episode to determine if the caregiver will respond to the infant. In a training run, the probability is fixed to be one of: 1\%, 5\%, 20\%, 80\%, 95\%, or 99\%. We refer to 95\% and 99\% as high-contingency (HC) and 1\% and 5\% as low-contingency (LC). 

We create validation sets from infants trained with HC and LC caregivers and test models trained with different levels of contingency on those sets. We decompose the world model loss into that due to infant orientation, position, and arm configuration ("Self"), Ball 1, Ball 2, and Caregiver predictions.

\subsection{Results}

\subsubsection{Increasing caregiver contingency shifts prediction difficulty from proprioceptive inputs to external dynamics}

The ``Self" component is harder to predict in validation cases with LC caregivers than with HC caregivers. This holds for both agents trained with LC and HC caregivers. In contrast, the presence of a HC caregiver corresponds with more difficult to predict ball and caregiver components (Figure \ref{disag-loss-by-contingency}). In the absence of very frequent caregiver interaction, agents focus on proprioceptive exploration. When present, the caregiver facilitates challenging external dynamics scenarios through its complex behavior patterns.

\subsubsection{The effect of high levels of contingency is asymmetric: world model performance improves substantially more on HC validation sets than it deteriorates on LC validation sets}
There is a decrease in error on the Ball 1, Ball 2, and Caregiver components on HC validation sets as the level of contingency increases. The caregiver agent facilitates challenging scenarios, which provide valuable experiences for the infant agents to learn from. In these same components, there is an increase in loss on LC validation sets with higher levels of caregiver contingency, but that increase is small on an absolute basis. This asymmetry appears to be persistent across training: it appears in early and late checkpoints (Figure \ref{disag-loss-by-contingency}).

The effect of increasing contingency on Self components appears nearly symmetric, possibly as a consequence of the previous result: the presence of a HC caregiver doesn't correspond to challenging examples, the difficulty decreases.

\section{Discussion and future work}

We find that basic social interaction and contingency activation can arise without requiring a specific module, social intrinsic reward, or extrinsic reward. It can arise from curiosity, implemented as an information-maximizing motivation.

An infant's ability to cause change in the world is amplified by an attentive caregiver, who responds to coos, cries, reaches and points. In our environment the caregiver is likely to facilitate the infant agent's first experience seeing a ball move. This amplification of change in the world is visible in our results: the prediction difficulty of external dynamics increases with a highly-contingent caregiver.

Contingent caregiver behavior provides a dense intrinsic reward for curiosity signals in a usually sparse environment. Social behavior is a very rich and difficult prediction problem, so each interaction will yield more data to challenge their understanding of the world. For signals that depend on a world model, this will strongly motivate exploration. Our results support this: multiple intrinsic reward functions frequently activate caregiver behaviors.

The social interaction that emerges when the infant agent is motivated by novelty (RND) or uncertainty (Disagreement) suggests the possibility that these classes of motivations could play a role in social behavior in children. The success of novelty is surprising: it suggests that a complex mechanism like a world model may not be needed for social behavior to emerge.  
Novelty alone is unlikely to explain exploratory behavior in infants \cite{gottlieb2013information}; another motivation is needed for infants to devote more attention to objects that violate expectations as seen in \citeA{stahl2015observing}. Uncertainty- or surprise-driven reward functions, could be more sensitive to situations where past experience would lead to multiple hypotheses or a different prediction that what occurred.
In our setup, these sensitivities do not appear necessary to yield good exploration and social interaction. Future work can investigate environment changes to elicit differences between novelty and uncertainty.

Adversarial rewards led the infant agent to repeatedly hit the ball during the Roll behavior because the model loss remained high. This is an instance of the white noise problem. Humans have mechanisms to avoid fixation on a single activity --- boredom, for example, has been considered as motivating a wider diversity of experience \cite{bench2013function}.

Future work could extend our simulation in important ways. The caregiver could engage in turn-taking similar to that seen in infant-caregiver pairs \cite{brazelton1974origins, gratier2015early}. The infant agent could use explicit representations of other agents' beliefs or model-based reinforcement learning to plan future actions.

In our simulation we can collect trajectories of the virtual infant agent. Real infant walking trajectories have been analyzed to understand exploration patterns and state coverage \cite{hoch2019s}. One important next step is to compare artificial and human trajectories on matched environments. This may lead to a better characterisation of infant exploratory motivations and patterns as well as insight into the observed diversity of exploration in children. 

\section{Acknowledgments}

The authors thank Isaac Kauvar, Julio Martinez, Logan Cross, Xi Jia Zhou, Eric Zelikman, and the entire Stanford Autonomous Agents Lab for their insightful discussions and feedback on the work. This work was supported by Stanford Human-Centered AI Hoffman-Yee (C.D. \& N.H.) and Google Cloud Credit Grants, a MediaX Learning Differences Taxonomies seed grant (C.D. \& N.H.), a Simons Foundation grant (543061, D.L.K.Y), a National Science Foundation CAREER grant (1844724,  D.L.K.Y), an Office of Naval Research grant (S5122, D.L.K.Y.), and the Stanford Graduate School of Education (C.D. \& N.H.).

\bibliographystyle{apacite}

\setlength{\bibleftmargin}{.125in}
\setlength{\bibindent}{-\bibleftmargin}

\bibliography{cd_20230116}

\end{document}